\documentclass{article}

\usepackage{graphicx} % more modern
\usepackage{subfigure} 

\usepackage{natbib}

\usepackage{algorithm}
\usepackage{algorithmic}

\usepackage{hyperref}

 \usepackage[accepted]{icml2013}
\icmltitlerunning{Dynamic Mirror Descent}
\usepackage{amsmath,amssymb,amsthm,graphicx,enumerate,url}
\usepackage{algorithm,algorithmic,subfigure,bm}
\def\htheta{{\widehat{\theta}}}
\def\ttheta{{\widetilde{\theta}}}

\def\argmin{\mathop{\!\arg \min}}
\def\argmax{\mathop{\!\arg \max}}
\def\deq{{\triangleq}}
\def\grad{\nabla}
\def\diam{D_{\rm max}}
\def\cV{{\nu}}

\def\btheta{\boldsymbol{\theta}}

\def\bx{\bm{x}}
\def\bhtheta{\widehat{\btheta}}

\newcommand{\wh}[1]{\widehat{#1}}
\newcommand{\wt}[1]{\widetilde{#1}}

\newcommand{\ave}[1]{\langle #1 \rangle}

\def\cV{{\cal V}}
\def\cL{{\cal L}}
\def\cF{{\cal F}}

\def\cR{{\cal R}}

\def\cX{{\mathsf X}}

\newtheorem{theorem}{{\bf Theorem}}

\newtheorem{definition}[theorem]{{\bf Definition}}
\newtheorem{lemma}[theorem]{{\bf Lemma}}

\newcommand{\ie}[0]{\emph{i.e., }}

\newcommand{\eg}[0]{\emph{e.g., }}

\newcommand{\bseq}{\begin{subequations}\begin{align}}
\newcommand{\eseq}{\end{align}\end{subequations}}

\begin{document} 

\twocolumn[
\icmltitle{Dynamical Models and Tracking Regret\\ in Online Convex Programming}

% unless you've provided the [accepted] option to the icml2013
\icmlauthor{Eric C. Hall}{ech11@duke.edu}
\icmlauthor{Rebecca M. Willett}{willett@duke.edu}
\icmladdress{Duke University, Department of Electrical and Computer Engineering,
            Durham, NC 27708}

\icmlkeywords{Online Optimization, Regularization, Stochastic Filtering, Dynamic Optimization, Sequential Prediction}

\vskip 0.3in
]

\begin{abstract}
  This paper describes a new online convex optimization method which
  incorporates a family of candidate dynamical models and establishes
  novel tracking regret bounds that scale with the comparator's deviation
  from the best dynamical model in this family. Previous online
  optimization methods are designed to have a total accumulated loss
  comparable to that of the best comparator sequence, and existing
  tracking or shifting regret bounds scale with the overall variation
  of the comparator sequence. In many practical scenarios, however,
  the environment is nonstationary and comparator sequences with small
  variation are quite weak, resulting in large losses. The proposed
  {\em Dynamic Mirror Descent} method, in contrast, can yield low
  regret relative to highly variable comparator sequences by both
  tracking the best dynamical model and forming predictions based on
  that model. This concept is demonstrated empirically in the context
  of sequential compressive observations of a dynamic scene and
  tracking a dynamic social network.
\end{abstract}

\section{Introduction}
In a variety of large-scale streaming data problems, ranging from
motion imagery formation to network analysis, dynamical models of the
environment play a key role in performance. Classical stochastic
filtering methods such as Kalman or particle filters or Bayesian
updates \cite{BaiCri09} readily exploit dynamical models for
effective prediction and tracking performance. However, classical
methods are also limited in their applicability because (a) they
typically assume an accurate, fully known dynamical model and (b) they
rely on strong assumptions regarding a generative model of the
observations. Some techniques have been proposed to learn the dynamics
\cite{XieSoh94,TheSha96}, but the underlying model still places heavy
restrictions on the nature of the data. Performance analysis of these
methods usually does not address the impact of ``model mismatch'',
where the generative models are incorrectly specified.

A contrasting class of prediction methods is based on an ``individual
sequence'' or ``universal prediction'' \cite{MerFed98} perspective;
these strive to perform provably well on any individual observation
sequence. In particular, online convex programming methods
\cite{NemYud83,BecTeb03,Zin03,CesLug06} rely on the gradient of the
instantaneous loss of a predictor to update the prediction for the
next data point.  The aim of these methods is to ensure that the
 per-round performance approaches that of the best {\em
  offline} method with access to the entire data sequence.  This
approach allows one to sidestep challenging issues associated with
statistically dependent or non-stochastic observations, misspecified
generative models, and corrupted observations. This framework is
limited as well, however, because performance bounds are typically
relative to either static or piecewise constant comparators and do not
adequately reflect adaptivity to a dynamic environment.

This paper describes a novel framework for prediction in the
individual sequence setting which incorporates dynamical models --
effectively a novel combination of state updating from stochastic
filter theory and online convex optimization from universal
prediction.  We establish tracking regret bounds for our proposed
algorithm, {\em Dynamic Mirror Descent (DMD),} which scale with the
deviation of a comparator sequence from a sequence evolving with a
known dynamic.  These bounds simplify to previously shown bounds, when
there are no dynamics.  We further establish tracking regret bounds
for another algorithm, {\em Dynamic Fixed Share (DFS)}, which
scale with the deviation of a comparator sequence from a sequence
evolving with the best {\em sequence} of dynamical models.  While our
methods and theory apply in a broad range of settings, we are
particularly interested in the setting where the dimensionality of the
parameter to be estimated is very high relative to the data volume.
In this regime, the incorporation of both dynamical models and
sparsity regularization plays a key role. With this in mind, we focus
on a class of methods which incorporate regularization as well as
dynamical modeling. The role of regularization, particularly sparsity
regularization, is increasingly well understood in batch settings and
has resulted in significant gains in ill-posed and data-starved
settings \cite{BanGhaAsp08,RavWaiLaf10,CS:noiseEC,BelNiy03}.

In our experiments, we consider reconstructing motion imagery from
sequential observations collected with a compressive camera and
estimating the dynamic social network underlying over 200 years of
U.S.\ Senate roll-call data.  There has been significant recent
interest in using models of temporal structure to improve time series
estimation from compressed sensing observations 
\cite{dynamicCS,modifiedCS} or for time-varying networks 
\cite{Sni01,xingNetwork}; the associated algorithms, however, are
typically batch methods poorly suited to large quantities of streaming
data. This paper strives to bridge that gap.

\section{Problem formulation}
\label{sec:prob}

\sloppypar Let $\cX$ denote the domain of our observations, and let
$\Theta$ denote a convex feasible set.  Given sequentially arriving
observations $\bm{x} \in \cX^\infty$, we wish to construct a sequence of
predictions $\bhtheta = (\htheta_1,\htheta_2,\ldots) \in
\Theta^\infty$, where $\htheta_{t}$ may depend only on the currently
available observations $\bx_{t-1} = (x_1,\ldots,x_{t-1})$. We pose our
problem as a dynamic game between a Forecaster and the Environment.
At time $t$, the Forecaster computes a prediction, $\htheta_{t}$ and
the Environment generates the observation $x_t$.  The Forecaster then
experiences the loss $\ell_t(\htheta_t)$, defined as follows.
Let $\cF$ and $\cR$ denote families of convex functions, and let
$f_t(\cdot) \deq f(\cdot,x_t) \in \cF$ be a cost function measuring
the accuracy of the prediction $\htheta_t$ with respect to the datum
$x_t$. Similarly, let $r(\cdot) \in \cR$ be a regularization term
which does not change over time; for instance, $r$ might promote
sparsity or other low-dimensional structure in the potentially
high-dimensional space $\Theta$. The loss 
at time $t$ is
$$\ell_t(\cdot) \deq f_t(\cdot) + r(\cdot)$$
where
$$\ell_t \in \cL \deq \{\ell = f + r: f \in \cF, r \in \cR\}.$$
The task facing the Forecaster is to create a new prediction
$\htheta_{t+1}$ based on the previous prediction and the new observation,
with the goal of minimizing loss at the next time step.  
We characterize the efficacy of $\wh{\btheta}_T \deq (\htheta_{1},
  \htheta_{2},\ldots,\htheta_T) \in \Theta^T$ relative to a
comparator sequence $\btheta_T\deq (\theta_1,\theta_2,\ldots,\theta_T) \in
\Theta^T$ as follows:

\begin{definition}[Regret] \label{defn:regret} The {\em regret} of
  $\bhtheta_T$ with respect to a comparator $\btheta_T \in \Theta^T$
  is
$$  R_T(\btheta_T) \deq  \sum_{t=1}^{T} \ell_{t}(\htheta_{t}) - \sum_{t=1}^{T}\ell_{t}(\theta_{t}).$$
\end{definition}

Previous work proposed algorithms which yielded regret of
$O(\sqrt{T})$ for {\em static} comparators, where $\theta_t = \theta$
for all $t$.  Our goal is to develop an online convex optimization
algorithm with low regret relative to a broad family of {\em
  time-varying} comparator sequences. In particular, our {\em main result}
is an algorithm which incorporates a dynamical model, denoted $\Phi_t$,
which admits a regret bound of the form $O(\sqrt{T}[1 + \sum_t \|
  \theta_{t+1} - \Phi_t( \theta_t)\|])$. This bound scales with the
  compartor sequence's deviation from the dynamical model $\Phi_t$ -- a
  stark contrast to previous tracking regret bounds which are only
  sublinear for comparators which change slowly with time or at a
  small number of distinct time instances.

%family of dynamical models, denoted $\Phi_1,\ldots,\Phi_N$, and
%results in a regret bound which scales like $O(\sqrt{T}[m\log N]

\section{Static, tracking, shifting, and adaptive regret}

In much of the online learning literature, the comparator sequence is
constrained to be static or time-invariant. In this paper we refer to
the regret with respect to a static comparator as {\em static regret}:
\begin{definition}[Static regret] \label{defn:static_regret} The {\em
    static regret} of $\bhtheta_T$ is
$$R_T \deq \sum_{t=1}^{T}\ell_t(\htheta_t) - \min_{\theta \in
    \Theta} \sum_{t=1}^{T} \ell_t(\theta).$$
\end{definition}

Static regret bounds are useful in characterizing how well an online
algorithm performs relative to, say, a loss-minimizing batch algorithm
with access to all the data simultaneously.  More generally, static
regret bounds compare the performance of the algorithm against a
static point, $\theta^*$, which can be chosen with full knowledge of
the data.

However, this form of analysis fails to illuminate the performance of
online algorithms in dynamic settings where a static comparator is
inappropriate.  Performance relative to a temporally-varying or
dynamic comparator sequence has been studied previously in the
literature in the context of tracking regret, shifting regret \cite{HerWar01,CesGaiLugSto12}, and the
closely-related concept of adaptive regret \cite{LitWar94,HazSes09}. 

In particular, tracking regret compares the output of the online
algorithm to a sequence of points
$\theta_1^*,\theta_2^*,...,\theta_T^*$ which can be chosen
collectively with full knowledge of the data.  This is a fair
comparison for a batch algorithm that detects and fits to drift in the
data, instead of fitting a single point.  Frequently,
in order to bound tracking regret there needs to be a measure of the
{\em complexity} of the sequence
$\theta_1^*,\theta_2^*,...,\theta_{T+1}^*$. Typically, this complexity is
characterized via a measure of the temporal variability of the
sequence, such as
$$V(\btheta_T) \deq \sum_{t=1}^{T} \|\theta_{t+1} - \theta_t\|.$$ If
this complexity is allowed to be very high, we could imagine that the
comparator series would fit the series of losses closely and hence
generalize poorly.  Conversely if this complexity is restricted to be
0, the tracking regret becomes equivalent to static regret.  Tracking
and shifting regret are the same concept, although the term shifting
regret is used more in the ``experts" setting, while tracking regret
tends to be a more generic term.

Adaptive regret is a related concept to tracking regret.  Instead of
measuring accumulated regret over the entire series, however, adaptive
regret measures accumulated loss over an arbitrary time interval
of length $\tau$, and measures performance against a static comparator chosen
optimally on this interval:
$$R_\tau \deq  \max_{[r, s] \subset [1, T]; s+1-r\leq \tau} 
\left [\sum_{t=r}^s \ell_t(\htheta_t) - \min_{\theta \in \Theta} \sum_{t=r}^s \ell_t(\theta)\right]$$
This is a valuable metric as it assures that a process will have low
loss not just globally, but also at any given moment.  Intuitively we
can see that an algorithm with low adaptive regret on any interval
should also have low tracking regret and vice versa. The relationship
between the two has been formally shown \cite{CesGaiLugSto12}.

In this paper, we present tracking/shifting regret bounds which rely
on a much more general notion of the complexity of a comparator
sequence. In particular, we could measure the complexity of a sequence
in terms of {\em how much it
deviates from a given dynamical model}, denoted $\Phi_t$:
\begin{align}
V_\Phi(\btheta_T) \deq \sum_{t=1}^{T} \| \theta_{t+1} -
\Phi_t(\theta_t)\|.
\label{eq:deviation}
\end{align} Ultimately, we consider a family of dynamical
models, and we measure the complexity of a comparator in terms of how
much it deviates from the best {\em sequence} of dynamical models in
this family.  (These concepts will be formalized and detailed in the
next two sections.)

It is intuitively satisfying that this measure appears in the bound.
Firstly, if the comparator actually follows the dynamics, we would
imagine this complexity to be very small, leading to low tracking regret.
This fact holds whether $\Phi_t$ is part of the
generative model for the observations or not. Secondly, we can get a dynamic
analog of static regret, where we enforce $V_\Phi(\btheta_T) = 0$.
This is equivalent to saying that the batch comparator is fitting the
best single trajectory using $\Phi_t$ instead of the best single point.
Using this, we would recover a bound analogous to a static regret bound
in a stationary setting.

Concurrent related work considers online algorithms where the data
sequence is described by a ``predictable process" \cite{Rak12}.  By
knowing a good estimate for the underlying process, they can create a
prediction sequence that follows accordingly, reducing 
overall loss.  However, they express their results in terms of a
static regret bound (\ie regret with respect to a static comparator)
with a variation term that expresses the deviation of the {\em input
  data} from the underlying process.  In contrast, we make no
assumptions about the data itself, but instead on the comparator
series, and form tracking regret bounds.

\section{Online convex optimization}
One common approach to
forming the predictions $\htheta_t$, Mirror Descent (MD)
\cite{NemYud83,BecTeb03}, consists of solving the following
optimization problem:
\begin{align}
\htheta_{t+1} = \argmin_{\theta \in \Theta} \eta_t \ave{\grad
  \ell_t(\htheta_t),\theta} + D(\theta \| \htheta_t),
\label{eq:md}
\end{align}
where $\grad \ell_t(\theta)$ denotes an arbitrary subgradient of
$\ell_t$ at $\theta$, $D(\theta \| \htheta_t)$ is the {\em Bregman
  divergence} between $\theta$ and $\htheta$, and $\eta_t \geq 0$ is a
step size parameter. 
Let $\psi$ denote a continuously differentiable function that is
$\sigma$-strongly convex with respect to a norm $\|\cdot\|$ on the set
$\Theta$ for some $\sigma > 0$; the Bregman divergence associated with
$\psi$ is defined as
\begin{subequations}
\begin{align}
D(\theta_1\| \theta_2)  =& D_\psi(\theta_1\|\theta_2)  \\
\deq&
\psi(\theta_1) - \psi(\theta_2) -
\ave{\nabla\psi(\theta_2),\theta_1-\theta_2} \\
\equiv& D(\theta_3\|\theta_2) +
D(\theta_1\|\theta_3) \nonumber \\ &+
\ave{\grad\psi(\theta_2)-\grad\psi(\theta_3),\theta_3-\theta_1}
\label{eq:triBreg}
\end{align}
\end{subequations}
for all $\theta_1,\theta_2,\theta_3 \in \Theta$,
and the strong convexity of $\psi$ implies
$$D(\theta_1\|\theta_2) \ge \frac{\sigma }{2}\|\theta_1-\theta_2\|^2.$$

The MD approach is a generalization of online learning algorithms such
as online gradient descent \cite{Zin03} and weighted majority
\cite{LitWar94}. Several recently proposed methods consider the
data-fit term separately from the regularization term
\cite{COMD,xiao,Lang09}. For instance, consider Composite Objective
Mirror Descent (COMD) \cite{COMD}:
\begin{align}
  \htheta_{t+1} = \argmin_{\theta \in \Theta} \eta_t
  \ave{\grad f_t(\htheta_t),\theta} + \eta_t r(\theta) + D(\theta \|
  \htheta_t).
\label{eq:comd}
\end{align}
This formulation is helpful when the regularization function
$r(\theta)$ promotes sparsity in $\theta$, and helps ensure that the
individual $\htheta_t$ are indeed sparse, rather than approximately
sparse as are the solutions to the MD formulation.  The regret of this
approach has previously been characterized as follows:
\begin{theorem}[Static regret for COMID \cite{COMD}]\label{thm:comid}
  Let $G_f \deq \max_{\theta \in \Theta, f \in \cF} \|\grad
  f(\theta)\|$, $\diam=\max_{\theta_1,\theta_2 \in \Theta} D(\theta_1\|\theta_2)$ and assume that $\theta \deq \theta_1 = \theta_2 =
  \cdots = \theta_T$. If $r(\htheta_1) = 0$ and $\eta_t = (2\sigma
  \diam)^{1/2}/(G_f \sqrt{T})$, then $$R_T(\btheta_T)
  \leq G_f(2T\diam/\sigma)^{1/2}.$$
\end{theorem}
\section{Dynamical models in online convex programming}
\label{sec:DMD}

Unlike the bound in Theorem~\ref{thm:comid},
{\em tracking} or {\em shifting regret}
\cite{CesLug06,CesGaiLugSto12} bounds typically consider
piecewise constant comparators, where $\theta_{t} - \theta_{t-1} = 0$
for all but $m$ values of $t$, where $m$ is a constant, or yield regret
bounds which scale with $\sum_t \| \theta_t - \theta_{t-1}\|$.  In
this paper, we develop tracking regret bounds which are small for much
broader classes of dynamic comparator sequences.

In particular, we propose the following alternative to \eqref{eq:md} and
\eqref{eq:comd}, which we call {\em Dynamic Mirror Descent (DMD)}. Let
$\Phi_t: \Theta \mapsto \Theta$ denote a predetermined dynamical model, and set
\begin{subequations}
\label{eq:dmd}
\begin{align}
\ttheta_{t+1} &= \argmin_{\theta \in \Theta} \eta_t
  \ave{\grad f_t(\htheta_t),\theta} + \eta_t r(\theta) + D(\theta \|
  \htheta_t)\label{eq:dmd1}\\
  \htheta_{t+1} &= \Phi_{t}( \ttheta_{t+1}) \label{eq:dmd2}
\end{align}
\end{subequations}
By including $\Phi_{t}$ in the process, we effectively search for a predictor which (a)
attempts to minimize the loss and (b) which is close to $\ttheta_t$
{\em under the transformation of $\Phi_t$.} 
This is similar to a stochastic filter which alternates between using
a dynamical model to update the ``state'', and then uses this state to
perform the filtering action. A key distinction of our approach,
however, is that we make no assumptions about $\Phi_t$'s relationship to
the observed data.

Our approach effectively includes dynamics into the COMID
approach. Indeed, for a case with no dynamics, so that $\Phi_t( \theta)
\equiv \theta$ for all $\theta$ and $t$, our method is equivalent to COMID.
Rather than considering COMID, we might have used other online
optimization algorithms, such as the Regularized Dual Averaging (RDA)
method \cite{xiao}, which has been shown to achieve similar
performance with more regularized solutions.  However, to the best of
our knowledge, no tracking or shifting regret bounds have been derived
for dual averaging methods (regularized or otherwise). Recent results
on the equivalence of COMID and RDA \cite{McMahan11} suggest that the
bounds derived here might also hold for a variant of RDA, but proving
this remains an open problem.

Our main result uses the following definitions:
\begin{align*}
G_\ell \deq& \max_{\theta \in \Theta, \ell \in \cL} \|\grad
\ell(\theta)\|\\
M \deq& \frac{1}{2}\max_{\theta \in
  \Theta}\|\nabla\psi(\theta)\|\\
\diam \deq& \max_{\theta, \theta' \in
  \Theta}D(\theta'\| \theta),\\
\mbox{and } \Delta_{\Phi_t} \deq&
\max_{\theta,\theta' \in \Theta}D(\Phi\theta\|\Phi\theta') -
D(\theta\|\theta'),
\end{align*}
\begin{theorem}
  \label{thm:main} Let $\Phi_t$ be a dynamical model such that
  $\Delta_{\Phi_t} \leq 0$. Let the sequence $\bhtheta_T$ be as in
  \eqref{eq:dmd2}, and let $\btheta_T$ be an arbitrary sequence in
  $\Theta^T$. Then the Dynamic Mirror Descent (DMD) algorithm using a
  non-increasing series $\eta_{t+1} \leq \eta_{t}$ gives 
\begin{align}
R_T(\btheta_T) \leq \frac{\diam}{\eta_{T+1}}& + \frac{4M}{\eta_T} V_{\Phi_t}(\btheta_T) +
\frac{G_{\ell}^2}{2\sigma} \sum_{t=1}^{T} \eta_t \\
 \mbox{with} \hspace{.1 in}
V_{\Phi_t}(\btheta_T) &\deq \sum_{t=1}^{T} \|\theta_{t+1} - \Phi_t(\theta_t)\| 
\end{align}
where $V_{\Phi_t}(\btheta_T)$
measures variations or deviations of the comparator sequence $\btheta_T$ from the
dynamical model $\Phi_t$.
\end{theorem}
Note that when $\Phi_t$ corresponds to an identity operator, the bound
in Theorem~\ref{thm:main} corresponds to existing tracking or shifting
regret bounds \cite{CesLug06,CesGaiLugSto12}.  The condition
that $\Delta_{\Phi_t} \leq 0$ is similar to requiring that $\Phi_t$ be a
contraction mapping.  This restriction is important; without it, any
poor prediction made at one time step could be magnified by
repeated application of the dynamics.  Additive models and matrix multiplications with 
all eigenvalues less than or equal to unity satisfy this restriction.  Notice also that if $\Phi_t=I$ for all $t$,
the theorem gives a novel tracking regret bound for COMID.  To prove
Theorem~\ref{thm:main}, we employ the following lemma, which is proven
in Section~\ref{pf:lem}.
\begin{lemma}\label{lem:main}
Let the sequence $\bhtheta_T$ be as in \eqref{eq:dmd2}, and let
$\btheta_T$ be an arbitrary sequence in $\Theta^T$; then
\begin{align*}
\ell_t(\htheta_t)-\ell_t(\theta_t) \leq
\frac{1}{\eta_t}\left[D(\theta_{t}\|\htheta_t) -
  D(\theta_{t+1}\|\htheta_{t+1}) \right]
\\ +\frac{\Delta_{\Phi_t}}{\eta_t}+
\frac{4M}{\eta_t}\|\theta_{t+1}-\Phi_t(\theta_{t})\| +
\frac{\eta_t}{2\sigma}G_\ell^2.
\end{align*}
\end{lemma}

\textbf{Proof of Theorem~\ref{thm:main}:} The proof is a matter of
summing the bounds of Lemma~\ref{lem:main} over time.  For simplicity denote
$D_t\deq D(\theta_{t}\|\htheta_{t})$ and $V_t \deq \|\theta_{t+1} -
\Phi_t(\theta_t)\|$. 
Then
\begin{align*}
R_T(\btheta_T) 
\le & \sum_{t=1}^{T}
\left(\frac{D_t}{\eta_{t}}-\frac{D_{t+1}}{\eta_{t+1}}\right) +  \frac{G_\ell^2}{2\sigma}\sum_{t=1}^{T} \eta_t
\\
& + D_{\max}\sum_{t=1}^{T}
\left(\frac{1}{\eta_{t+1}}-\frac{1}{\eta_t}\right)  + \sum_{t=1}^{T}
\frac{4M}{\eta_t}V_t
\\
\le & \frac{D_{\max}}{\eta_{T+1}} +
\frac{4M}{\eta_T}V_{\Phi_t}(\btheta_T) +
\frac{G_\ell^2}{2\sigma}\sum_{t=1}^{T} \eta_t. & \Box
\end{align*}

We set $\eta_t$ using the doubling trick \cite{CesLug06} whereby 
time is divided into increasingly longer segments, and on each
interval a temporary time horizon is fixed, known, and used to
determine an optimal step size (generally proportional to the inverse
of the square root of the time
horizon). 
This approach yields the regret bound:
\begin{align*}
R(\btheta_T) = O(\sqrt{T}[1+V_{\Phi_t}(\btheta_T)])
\end{align*}
This proof shares some ideas with the tracking regret bounds of \cite{Zin03}, but uses properties of the Bregman Divergence to eliminate some terms, while additionally incorporating dynamics.

%%%%%%%%%%%%%%%%%%%Assortment of Dynamics, Static Bound%%%%%%%%%%
\section{Prediction with a family of dynamical models}
\label{sec:PWEA}
DMD in the previous section uses a single dynamical model. In practice,
however, we do not know the best dynamical model to use, or the best
model may change over time in nonstationary environments.

To address this challenge, we assume a finite set of candidate
dynamical models $\{\Phi^{(1)}_t, \Phi^{(2)}_t, \ldots \Phi^{(N)}_t\}$, and describe a
procedure which uses this collection to adapt to nonstationarities in
the environment.  In particular, we establish tracking regret bounds
for a comparator class with {\em different dynamical models on different
time intervals}.  This class, $\Theta_m$, can be described as all
predictors defined on $m+1$ segments $[t_i,t_{i+1}-1]$ with time points $1= t_1 <
\cdots < t_{m+2} = T+1$. For a given $\btheta_T \in \Theta_m$ and $k =
1,\ldots,m+1$, let
\begin{align*}V&^{(m+1)}(\btheta_T) \deq\\
&\min_{t_2, \ldots, t_{m+1}}
\sum_{k=1}^{m+1} \min_{i_k \in \{1,\ldots,N\}} \sum_{t=t_k}^{t_{k+1}-1} \|
\theta_{t+1}-\Phi^{(i_k)}_t(\theta_t) \|
\end{align*}
%\wt{V}_k(\btheta_T) \deq \min_i \sum_{t=t_k}^{t_{k+1}-1} \|\Phi_{i}
%\theta_t - \theta_{t+1}\|$$ 
denote the deviation of the sequence
$\btheta_T$ from the best series of $m+1$ dynamical models.

Let $\htheta_t^{(i)}$ denote the output of the DMD algorithm of
Section~\ref{sec:DMD} using dynamical model $\Phi^{(i)}_t$. Then tracking
regret can be expressed as:
\begin{align} R_T(\Theta_m) =& \underbrace{\sum_{t=1}^{T} \ell_t(\htheta_t) -
  \min_{i_1,\ldots,i_T} \sum_{t=1}^{T}
  \ell_t\left(\htheta_t^{(i_t)}\right)}_{T_1} \nonumber\\ 
+& \underbrace{\min_{i_1,\ldots,i_T} \sum_{t=1}^{T}
  \ell_t\left(\htheta_t^{(i_t)}\right) - \min_{\btheta \in \Theta_m}
  \sum_{t=1}^{T} \ell_t(\theta_t)}_{T_2} \label{Tracking_Decomp}
\end{align}
where the minimization in the second term of $T_1$ and first term of
$T_2$ is with respect to sequences of dynamical models with at most
$m$ switches, such that $\sum_{t=1}^T \mathbf{1}_{[i_t \neq
    i_{t+1}]}\leq m$.  In \eqref{Tracking_Decomp}, $T_1$ corresponds
to the tracking regret of our algorithm relative to the best sequence
of dynamical models within the DMD framework, and $T_2$ is the regret
of that sequence relative to the best comparator in the class
$\Theta_m$.

We choose $\htheta_t$ by using the Fixed Share (FS) forecaster on the
DMD estimates of \eqref{eq:dmd}, $\htheta_t^{(i)}$. In FS, each expert (here, each
candidate dynamical model) is assigned a weight that is inversely
proportional to its cumulative loss at that point yet with some weight
shared amongst all the experts, so that an expert with very small
weight can quickly regain weight to become the leader
\cite{CesLug06}. Our estimate is:
\begin{align}
\wt{w}_{i,t}=&w_{i,t-1}\exp(-\eta_r \ell_t(\htheta_t^{(i)})) \\
w_{i,t}=&(\lambda/N) \textstyle\sum_{j=1}^N \wt{w}_{j,t} + (1-\lambda)
\wt{w}_{i,t} \\
\htheta_{t}=&\sum_{i=1}^N w_{i,t}\htheta_t^{(i)} / \sum_{i=1}^N w_{i,t}. \label{RFS}
\end{align}

Following \cite{CesLug06}, we have
\begin{align}
T_1 \leq \frac{m+1}{\eta_r} \log N+ \frac{1}{\eta_r}
\log \frac{1}{\lambda^m (1-\lambda)^{(T-m-1)}} +
\frac{\eta_r}{8}T %\label{Tracking2}%
\nonumber
\end{align}
and $T_2$ can be bounded using the method described in
Section~\ref{sec:DMD} on each time interval $[t_k,t_{k+1}-1]$ and
summing over the $m+1$ intervals, yielding 
\begin{align}
T_2 \leq \frac{(m+1)\diam}{\eta_{T+1}} + \frac{4M}{\eta_T}V^{(m+1)}(\btheta_T) + \frac{G_{\ell}^2}{2\sigma} \sum_{t=1}^{T}
\eta_t. %\label{Tracking1}
\nonumber
\end{align}
Letting $\eta_r = \eta_t = 1/\sqrt{T}$, the overall expected tracking regret is thus
\begin{align*}
R_T(&\btheta_T) = O\Big(\sqrt{T} \Big[(m+1)(\log N+\diam)\\
&+    \log \frac{1}{\lambda^m (1-\lambda)^{(T-m-1)}} + 4M
  V^{(m+1)}(\btheta_T) \Big]\Big).
\end{align*}
The last term in this bound measures the
deviation of a comparator in $\Theta_m$ from the best series of
dynamical models over $m+1$ segments (where $m$ does not scale with $T$).  Here $\lambda$ is usually chosen to
be $\frac{m}{T}$ where $m$ is an upper bound on the number of
switches, independent of $T$. Again, if $T$ is not known in advance
the doubling trick can be used.  Note that $
V^{(m+1)}(\btheta_T) \leq V_{\Phi_t^{(i)}}(\btheta_T)$ for any fixed $i \in
\{1,\ldots,N\}$, thus this approach generally yields lower regret than
using a fixed dynamical model.  However, we incur some loss by not
knowing the optimal number of switches $m$ or when the switching times
are; these are accounted for in $T_1$.

We use the Fixed Share algorithm as a means to amalgamate estimates with different dynamics, however
other methods could be used with various tradeoffs.  The Fixed Share algorithm, for instance, has linear complexity
with low regret, but with respect to a comparator class with fixed number of switches.  Other algorithms can accommodate larger
classes of experts, or not assume knowledge of the number of switches, but come at the price of higher regret or complexity as explained in \cite{Gyo12}.

\section{Experiments and results}
To demonstrate the performance of Dynamic Mirror Descent (DMD) combined with 
the Fixed share algorithm (which we call {\em Dynamic Fixed Share (DFS)}), we
consider two scenarios: reconstruction of a dynamic scene (\ie video)
from sequential compressed sensing observations, and tracking 
connections in a dynamic social network.
\subsection{Compressive video reconstruction}
To test DMD, we construct a video which contains an object 
moving in a 2-dimensional plane; the $t^{\rm th}$ frame is denoted
$\theta_t$ (a $150 \times 150$ image stored as a length-$22500$
vector) which takes values between 0 and 1.
The corresponding observation is $x_t = A_t\theta_t + n_t,$
where $A_t$ is a random $500 \times 22500$ matrix and $n_t$
corresponds to measurement noise. This model coincides with several
compressed sensing architectures \cite{riceCamera}. 
We used white Gaussian noise with variance 1.

Our loss function uses
$f_t(\theta) = \frac{1}{2}\|x_t - A_t\theta\|_2^2$ and $r(\theta) = \tau\|\theta\|_1$,
where $\tau > 0$ is a tuning parameter.
We construct a family of $N = 9$ dynamical models, where
$\Phi_t^{(i)} (\theta)$ shifts the frame, $\theta$, one pixel in a direction
corresponding to an angle of $2\pi i/(N-1)$ as well as a ``dynamic''
corresponding to no motion. (With the static model, DMD reduces to COMID.) The true video sequence uses
different dynamical models over $t=\{1,...,240\}$ and $t= \{241,...,500\}$.
Finally, we use $\psi(\cdot)
= \|\cdot\|^2_2$ so the Bregman Divergence $D(x\|y)=\|x-y\|^2_2$ is
the usual squared Euclidean distance. The DFS forecaster uses $\lambda=0.01$.

\begin{figure}[h]
\centering
\includegraphics[width=2.25 in, height=2.25 in]{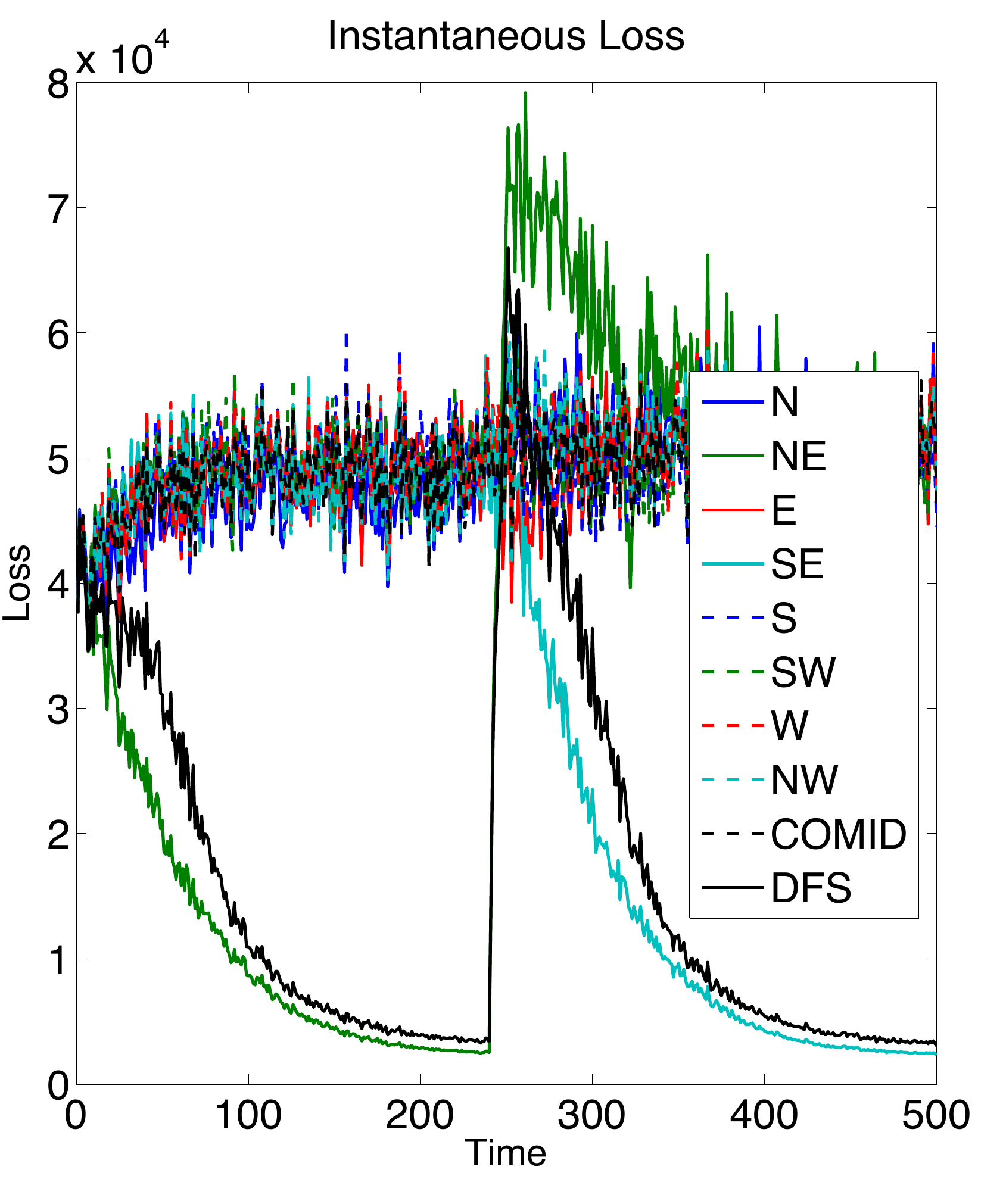}
\caption{Tracking dynamics using DFS and comparing individual models for directional (N, S, E, etc.) motion.
  Before $t=240$ the NE motion dynamic model
incurs small loss, where as after $t=240$ the SE motion does well, and DFS
successfully tracks this change.}
\label{fig:cs1}
\end{figure}

\begin{figure}[t]
\centering
\includegraphics[width=2.5 in, height=2.7 in]{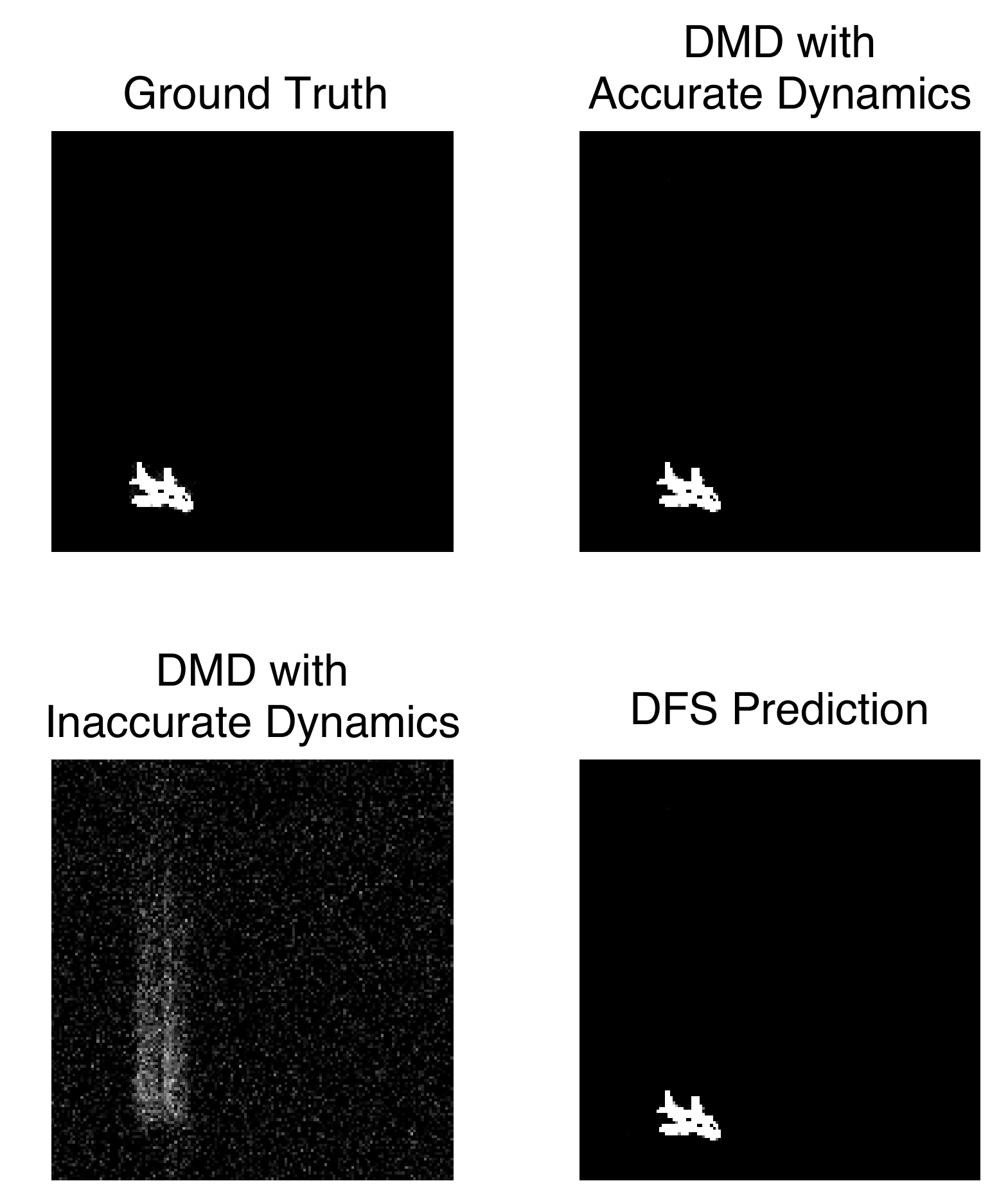}
\caption{Instantaneous predictions at $t=480$.  Top Left: $\theta_t$.  Top Right: $\htheta^{(SE)}_t$.  
Bottom Left: $\htheta_t^{(E)}$.  Bottom Right: $\htheta_t$.  The prediction made with the
prevailing motion is an accurate representation of the ground truth, while the prediction with the wrong dynamic is 
an unclear picture.  The DFS algorithm correctly picks out the cleaner picture.}
\label{fig:cs2}
\end{figure}

Figures \ref{fig:cs1} and \ref{fig:cs2} show the impact of using DFS.
We see that DFS switches between dynamical models rapidly and outperforms all of the individual
predictions, including COMID, used as a baseline, to show the advantages of incorporating knowledge of the dynamics.

\subsection{Tracking dynamic social networks}
Dynamical models have a rich history in the context of social network
analysis \cite{Sni01}, but we are unaware of their application in the
context of online learning algorithms. To show how DMD can bridge this
gap, we track the influence matrix of seats in the US Senate from 1795
to 2011 using roll call data (http://www.voteview.com/dwnl.htm). At
time $t$, we observe the ``yea'' or ``nay'' vote of each Senator,
which we represent with a $+1$ or $-1$. When a Senator's vote is
unavailable (for instance, before a state joined the union), we use a
$0$. We form a length $p=100$ vector of these votes indexed by the
Senate seat, and denote this $x_t$.

Following \cite{RavWaiLaf10}, we form a loss function using a negative
log Ising model {\em pseudolikelihood} to sidestep challenging issues
associated with the partition function of the Ising model likelihood.  
For a social network with $p$ agents, $\theta_t \in [-1,1]^{p \times
  p}$, where $(\theta_t)_{ab}$ corresponds to the correlation in voting
patterns between agents $a$ and $b$ at time $t$. Let $\cV$ denote the set of
agents, $\cV \backslash a$ the set of all agents except $a$, $x_a$ the
vote of agent $a$, and $\theta_ a \deq \{\theta_{ a b}: b \in
\cV\}$. Our loss function is
\begin{subequations}
\begin{align*}
\varphi^{( a)}_{t}( \theta_ a) \deq& \log\left[\exp\left(2\theta_{ a a
    }x_ a + 2 \textstyle\sum_{ b \in \cV \backslash 
      a}\theta_{ a b}x_ a x_ b  \right) + 1\right]\\
f^{(a)}(\theta_a;x) \deq& -2\theta_{ a a }x_ a - 2 \textstyle\sum_{ b
  \in \cV\backslash a}\theta_{ a b}x_ a x_ b  + \varphi^{( a)}_{t}(
\theta_ a)\\ 
f(\theta; x) =& \textstyle\sum_{ a \in \cV} f^{(a)}(\theta_a;x)
%\label{eq:cond}
\end{align*}
\end{subequations}
and $r(\theta) = \tau\|\theta\|_1$, where $\tau > 0$ is a tuning
parameter; this loss is convex in $\theta$. We set $\psi(\theta) =
\frac{1}{2}\|\theta\|^2_2$ and
use a dynamical model inspired by \cite{Sni01}, where if $|\theta_{ac^*}\theta_{bc^*}| > |\theta_{ab}|$, with $c^* = \argmax_c |\theta_{ac}\theta_{bc}|$, then: 
$$
(\Phi_t^{(i)}(\theta))_{ab}=(1-\alpha_i) \theta_{ab} +
  \alpha_i \theta_{ac^*}\theta_{bc^*}.
$$
%\begin{gather}
%(\Phi_t^{(i)} (\theta))_{ab} = \nonumber \\
%\begin{cases} (1-\alpha_i) \theta_{ab} +
%  \alpha_i \theta_{ac^*}\theta_{bc^*} & \mbox{if }
%    |\theta_{ac^*}\theta_{bc^*}| > |\theta_{ab}| \\
%\theta_{ab} & \mbox{otherwise} \end{cases}, \nonumber
%\end{gather}
Otherwise, $\htheta^{(i)}_{ab}=\ttheta^{(i)}_{ab}$. The intuition is that if two members of the network share a strong
common connection, they will become connected in time.  We set
$\alpha_i \in \{0,.001,.002,.003,.004\}$ for the different dynamical
models.  We set $\tau = .1$ and again set $\eta$ using the doubling
trick with time horizons at set at increasing powers of 10.  As in
\cite{Lang09}, we find that regularizing (\eg thresholding) every 10
steps, instead of at each time step, allows for the values to grow
above the threshold for meaningful relationships to be found.

\begin{figure}[H]
\centering
\includegraphics[height=2.2 in, width=3  in]{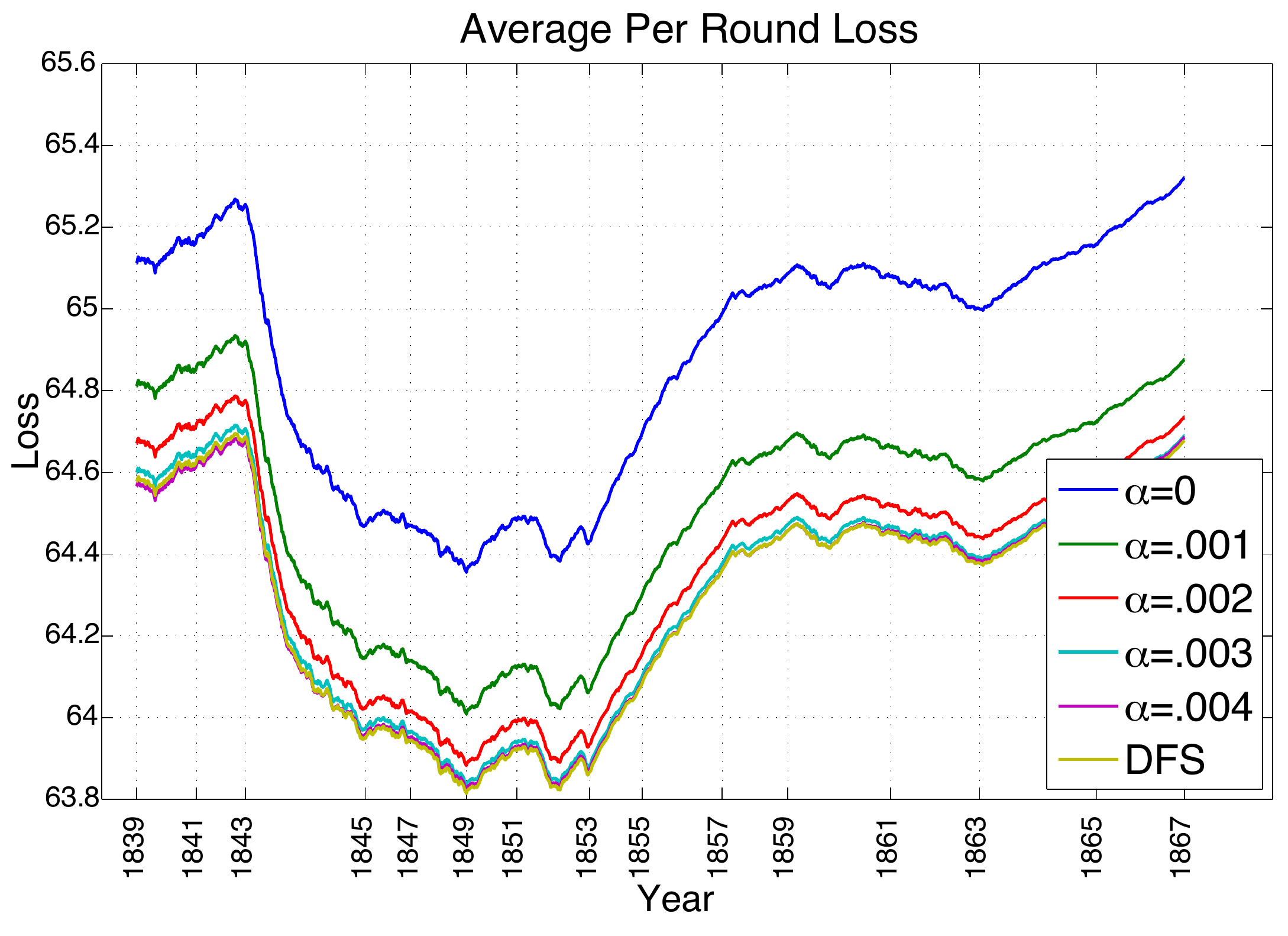}
\caption{Tracking a dynamic social network. Losses for
  different dynamical models and the DFS predictions; $\alpha = 0$
  corresponds to COMID.}
\label{fig:senate}
\end{figure}

\begin{figure}[H]
\centering
\includegraphics[width=3 in, height=2.2 in]{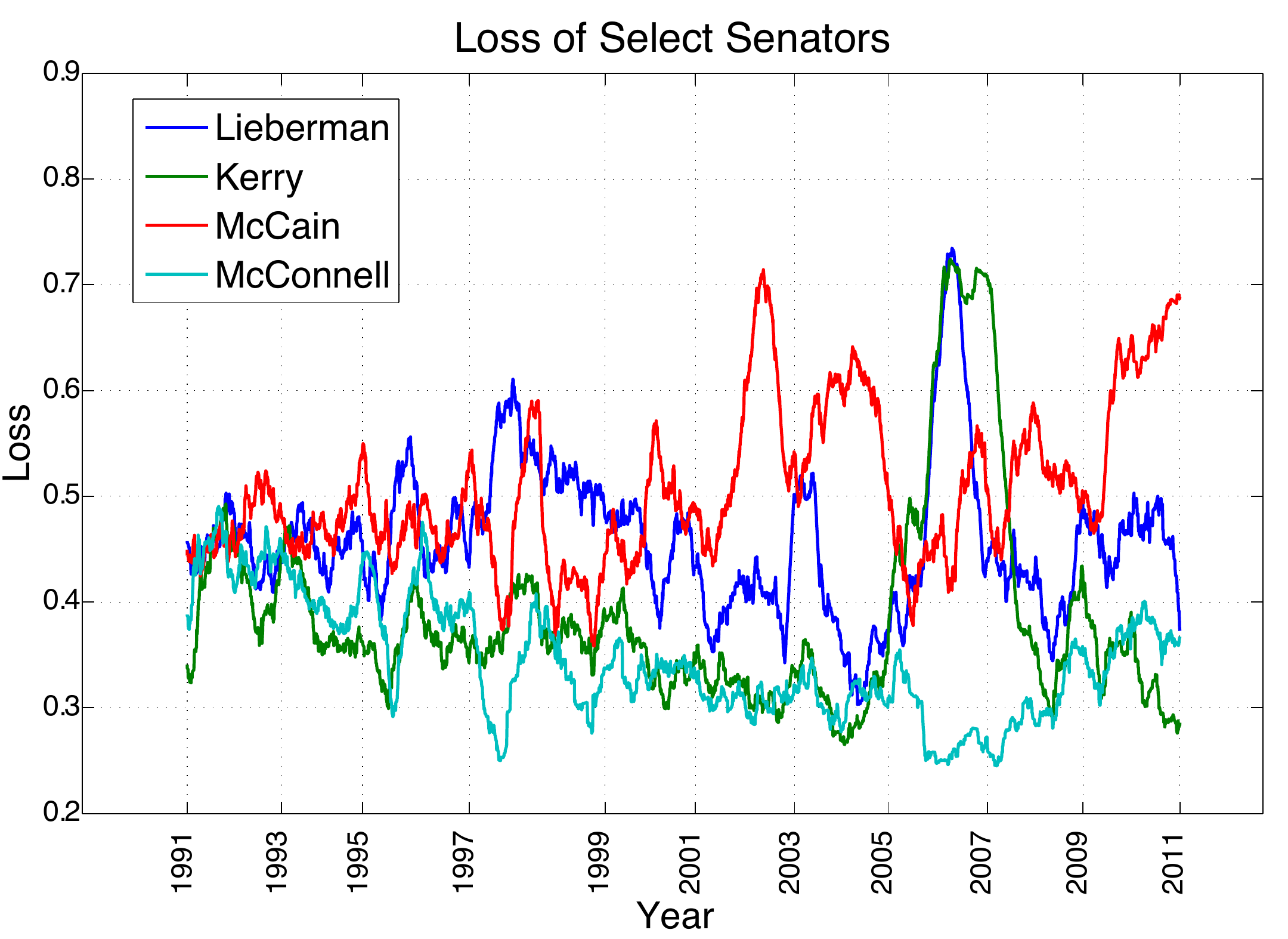}
\caption{Losses for individual senators. Low losses correspond to 
predictable, consistent voting behavior, while higher loss means less predictable}
\label{fig:sen_local}
\end{figure}

Figure \ref{fig:senate} shows the average per round
loss of each model, and the DFS estimator over a 30 year time window.
We see that applying the dynamical model improves
performance relative to COMID ($\alpha_i = 0$) and that DFS
aggregates the predictions successfully.  Figure \ref{fig:sen_local}
shows the moving average losses for a few Senators, where high
loss corresponds to behavior unexpected in the model.  Notice that John Kerry
(D-MA) has generally low loss, spikes around 2006, and then
drops again before a reelection campaign in 2008.

\begin{figure}[H]
\centering
\includegraphics[width=3 in, height=2.00 in]{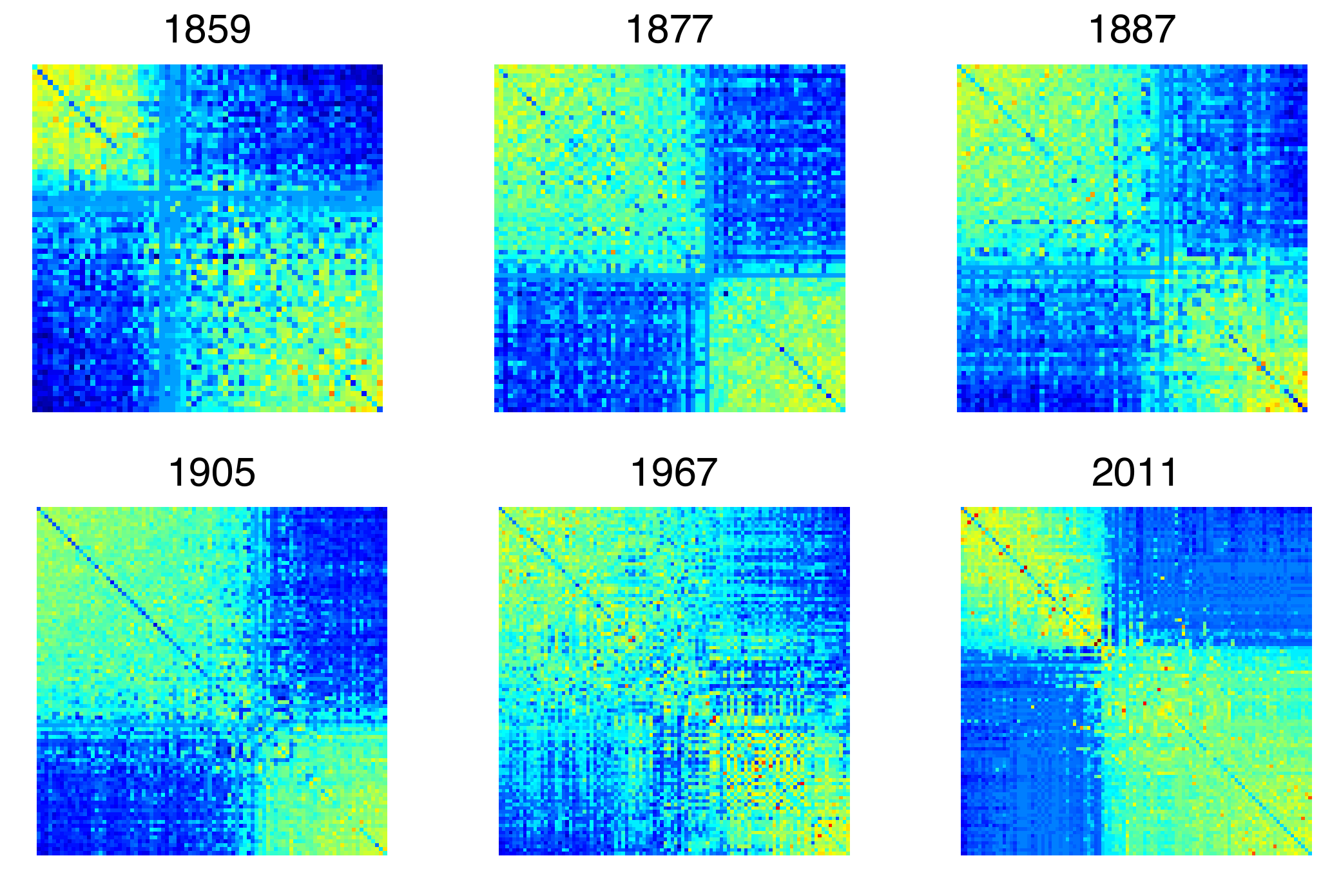}
\caption{Influence matrices for
  select years spanning Civil War and Civil Rights Movement to
  present, showing formation of factions. 
  Warmer colors (reds and greens) correspond to higher influence and colder (blue) corresponds to 
  lower influence. (Best viewed in color)}
\label{fig:adjmats}
\end{figure}

Looking at the network estimates of DFS across time
(as in Figure~\ref{fig:adjmats}) we can see tight factions forming in the mid- to late-1800s (post Civil War), 
followed by a time when the factions dissipate in the mid-1900s during
the Civil Rights Movement.  Finally, we see factions again forming in more recent
times.  The seats are sorted separately for each matrix to emphasize
groupings, which align with known political
factions.

\section{Conclusion and future directions}
In this paper we have proposed a novel online optimization method,
called Dynamic Mirror Descent (DMD), which incorporates dynamical
model state updates. There is no assumption that there is a ``true''
known underlying dynamical model, or that the best dynamical model is
unchanging with time. The proposed Dynamic Fixed Share (DFS) algorithm
adaptively selects the most promising dynamical model from a family of
candidates at each time step.  Recent work on shifting or tracking
regret bounds for online convex optimization further suggest that the
techniques developed in this paper may also be useful for bounding
adaptive regret or developing methods for automatically tuning
step-size parameters \cite{CesGaiLugSto12}.  In experiments with real
and simulated data, DMD shows strong tracking behavior even when
underlying dynamical models are switching.
\section{Proofs}
\label{pf:lem}
{\bf Proof of Lemma~\ref{lem:main}:}

The optimality condition of \eqref{eq:dmd1} implies
\begin{align}
\ave{\grad f_t(&\htheta_t)+\grad
  r(\ttheta_{t+1}),\ttheta_{t+1}-\theta_{t}} \leq \nonumber
\\ &\frac{1}{\eta_t} \ave{\grad \psi(\htheta_t) -
  \grad \psi(\ttheta_{t+1}), \ttheta_{t+1}-\theta_{t}}.
\label{eq:firstord}
\end{align}
Using this condition we can bound the instantaneous regret as follows:
\begin{subequations}
\begin{align}
f_t&(\htheta_t)-f_t(\theta_t)+r(\htheta_{t})-r(\theta_{t}) \nonumber\\
= &f_t(\htheta_t)-f_t(\theta_t)+r(\htheta_{t}) - r(\ttheta_{t+1}) +r(\ttheta_{t+1})-r(\theta_{t})\nonumber \\
\leq&  \ave{ \grad f_t( \htheta_t), \htheta_t - \theta_t } 
+ \ave{ \grad r(\htheta_{t}),\htheta_{t}-\ttheta_{t+1} } \nonumber\\
&+ \ave { \grad r(\ttheta_{t+1}),\ttheta_{t+1} - \theta_t } \label{eq:useConvexity}\\
\le& \frac{1}{\eta_t} \ave{\grad \psi(\htheta_t) - \grad \psi (\ttheta_{t+1}), \ttheta_{t+1}-\theta_{t}} \nonumber \\
 &+ \ave{\grad f_t (\htheta_{t}) + \grad r (\htheta_{t}),  \htheta_t -\ttheta_{t+1}}\label{eq:useFirst}\\
 =& \frac{1}{\eta_t} \left[D(\theta_{t}\|\htheta_t) -
  D(\theta_{t+1}\|\htheta_{t+1})\right] \nonumber \\
&+T_3 + T_4/\eta_t + T_5/\eta_t \label{eq:useTriBreg2} \hspace{.25 in} \text{where,}\\
T_3 \deq& - \frac{1}{\eta_t} D(\ttheta_{t+1}\|\htheta_t)
 + \ave{\grad \ell_t(\htheta_t),\htheta_t-\ttheta_{t+1}} \nonumber\\
T_4 \deq & \left[
  D(\Phi_t(\theta_{t})\|\Phi_t(\ttheta_{t+1}))  -
  D(\theta_{t}\|\ttheta_{t+1}) \right] \leq \Delta_{\Phi_t} \nonumber\\
T_5 \deq&  \left[  D(\theta_{t+1}\|\htheta_{t+1}) -
  D(\Phi_t(\theta_{t})\|\htheta_{t+1})\right]. \nonumber
\end{align}
\end{subequations}
Here, \eqref{eq:useConvexity} follows from the convexity of $f_t$ and
$r$, \eqref{eq:useFirst} follows from the optimality condition of \eqref{eq:dmd1}, and
\eqref{eq:useTriBreg2} follows from \eqref{eq:triBreg} and adding and subtracting terms using the equivalence \eqref{eq:dmd2}.
Each of term can be bounded, and then combined to complete the proof.
\begin{subequations}
\begin{align}
T_3  \leq& -\frac{\sigma}{2\eta_t} \| \ttheta_{t+1}-\htheta_t\|^2\nonumber \\
 &+\frac{\sigma}{2\eta_t}\|\ttheta_{t+1}-\htheta_t\|^2 +
\frac{\eta_t}{2\sigma}G_\ell^2 \label{eq:young}
=\frac{\eta_t}{2\sigma}G_\ell^2\\
T_5  =&  \psi(\theta_{t+1})  - \ave{ \grad
\psi(\htheta_{t+1}),\theta_{t+1}-\htheta_{t+1} } \nonumber\\
&-\psi(\Phi_t(\theta_{t})) + \ave{\grad
  \psi(\htheta_{t+1}),\Phi_t(\theta_{t}) - \htheta_{t+1}}\nonumber \\
=& \psi(\theta_{t+1}) -\psi(\Phi_t(\theta_{t})) - \ave{ \grad
\psi(\htheta_{t+1}),\theta_{t+1}-\Phi_t(\theta_t)} \nonumber \\
\leq&  4M\|\theta_{t+1} - \Phi_t(\theta_{t})\|\label{eq:Cauch}
\end{align}
\end{subequations}
where \eqref{eq:young} is due to the strong convexity of the Bregman Divergence and 
Young's inequality and \eqref{eq:Cauch} is due to the
convexity of $\psi$ and the Cauchy-Schwarz inequality. Combining these
inequalities with \eqref{eq:useTriBreg2} gives the Lemma as it is
stated. \hfill $\Box$

\newpage
\bibliography{DynamicOCP}

\begin{thebibliography}{27}
\providecommand{\natexlab}[1]{#1}
\providecommand{\url}[1]{\texttt{#1}}
\expandafter\ifx\csname urlstyle\endcsname\relax
  \providecommand{\doi}[1]{doi: #1}\else
  \providecommand{\doi}{doi: \begingroup \urlstyle{rm}\Url}\fi

\bibitem[Angelosante et~al.(2009)Angelosante, Giannakis, and Grossi]{dynamicCS}
Angelosante, D., Giannakis, G.~B., and Grossi, E.
\newblock Compressed sensing of time-varying signals.
\newblock In \emph{Int’l Conf. on Dig. Sig. Proc.}, 2009.

\bibitem[Bain \& Crisan(2009)Bain and Crisan]{BaiCri09}
Bain, A. and Crisan, D.
\newblock \emph{Fundamentals of Stochastic Filtering}.
\newblock Springer, 2009.

\bibitem[Banerjee et~al.(2008)Banerjee, {El Ghaoui}, and
  {d'Aspremont}]{BanGhaAsp08}
Banerjee, O., {El Ghaoui}, L., and {d'Aspremont}, A.
\newblock Model selection through sparse maximum likelihood estimation for
  multivariate {G}aussian or binary data.
\newblock \emph{J. Mach. Learn. Res.}, 9:\penalty0 485--516, 2008.

\bibitem[Beck \& Teboulle(2003)Beck and Teboulle]{BecTeb03}
Beck, A. and Teboulle, M.
\newblock Mirror descent and nonlinear projected subgradient methods for convex
  programming.
\newblock \emph{Operations Research Letters}, 31:\penalty0 167--175, 2003.

\bibitem[Belkin \& Niyogi(2003)Belkin and Niyogi]{BelNiy03}
Belkin, M. and Niyogi, P.
\newblock Laplacian eigenmaps for dimensionality reduction and data
  representation.
\newblock \emph{Neural Comput.}, 15\penalty0 (6):\penalty0 1373--1396, June
  2003.

\bibitem[Cand\`{e}s et~al.(2006)Cand\`{e}s, Romberg, and Tao]{CS:noiseEC}
Cand\`{e}s, E., Romberg, J., and Tao, T.
\newblock Stable signal recovery from incomplete and inaccurate measurements.
\newblock \emph{Communications on Pure and Applied Mathematics}, 59\penalty0
  (8):\penalty0 1207--1223, 2006.

\bibitem[Cesa-Bianchi \& Lugosi(2006)Cesa-Bianchi and Lugosi]{CesLug06}
Cesa-Bianchi, N. and Lugosi, G.
\newblock \emph{Prediction, Learning and Games}.
\newblock Cambridge University Press, New York, 2006.

\bibitem[Cesa-Bianchi et~al.(2012)Cesa-Bianchi, Gaillard, Lugosi, and
  Stoltz]{CesGaiLugSto12}
Cesa-Bianchi, N., Gaillard, P., Lugosi, G., and Stoltz, G.
\newblock A new look at shifting regret.
\newblock arXiv:1202.3323, 2012.

\bibitem[Duarte et~al.(2008)Duarte, Davenport, Takhar, Laska, Sun, Kelly, and
  Baraniuk]{riceCamera}
Duarte, M.~F., Davenport, M.~A., Takhar, D., Laska, J.~N., Sun, T., Kelly,
  K.~F., and Baraniuk, R.~G.
\newblock Single pixel imaging via compressive sampling.
\newblock \emph{IEEE Sig. Proc. Mag.}, 25\penalty0 (2):\penalty0 83--91, 2008.

\bibitem[Duchi et~al.(2010)Duchi, Shalev-Shwartz, Singer, and Tewari]{COMD}
Duchi, J., Shalev-Shwartz, S., Singer, Y., and Tewari, A.
\newblock Composite objective mirror descent.
\newblock In \emph{Conf. on Learning Theory (COLT)}, 2010.

\bibitem[Gyorgy et~al.(2012)Gyorgy, Linder, and Lugosi]{Gyo12}
Gyorgy, A., Linder, T., and Lugosi, G.
\newblock Efficient tracking of large classes of experts.
\newblock \emph{IEEE Transaction on Information Theory}, 58:\penalty0
  6709--6725, November 2012.

\bibitem[Hazan \& Seshadhri(2009)Hazan and Seshadhri]{HazSes09}
Hazan, E. and Seshadhri, C.
\newblock Efficient learning algorithms for changing environments.
\newblock In \emph{Proc. Int. Conf on Machine Learning (ICML)}, pp.\  393--400,
  2009.

\bibitem[Herbster \& Warmuth(2001)Herbster and Warmuth]{HerWar01}
Herbster, M. and Warmuth, M.~K.
\newblock Tracking the best linear predictor.
\newblock \emph{Journal of Machine Learning Research}, 35\penalty0
  (3):\penalty0 281--309, 2001.

\bibitem[Kolar et~al.(2010)Kolar, Song, Ahmed, and Xing]{xingNetwork}
Kolar, M., Song, L., Ahmed, A., and Xing, E.~P.
\newblock Estimating time-varying networks.
\newblock \emph{Annals of Applied Statistics}, 4\penalty0 (1):\penalty0
  94--123, 2010.

\bibitem[Langford et~al.(2009)Langford, Li, and Zhang]{Lang09}
Langford, J., Li, L., and Zhang, T.
\newblock Sparse online learning via truncated gradient.
\newblock \emph{J. Mach. Learn. Res.}, 10:\penalty0 777--801, 2009.

\bibitem[Littlestone \& Warmuth(1994)Littlestone and Warmuth]{LitWar94}
Littlestone, N. and Warmuth, M.~K.
\newblock The weighted majority algorithm.
\newblock \emph{Inf. Comput.}, 108\penalty0 (2):\penalty0 212--261, 1994.

\bibitem[McMahan(2011)]{McMahan11}
McMahan, B.
\newblock A unified view of regularized dual averaging and mirror descent with
  implicit updates.
\newblock arXiv:1009.3240v2, 2011.

\bibitem[Merhav \& Feder(1998)Merhav and Feder]{MerFed98}
Merhav, N. and Feder, M.
\newblock Universal prediction.
\newblock \emph{IEEE Trans. Info. Th.}, 44\penalty0 (6):\penalty0 2124--2147,
  October 1998.

\bibitem[Nemirovsky \& Yudin(1983)Nemirovsky and Yudin]{NemYud83}
Nemirovsky, A.~S. and Yudin, D.~B.
\newblock \emph{Problem complexity and method efficiency in optimization}.
\newblock John Wiley \& Sons, New York, 1983.

\bibitem[Rakhlin \& Sridharan(2012)Rakhlin and Sridharan]{Rak12}
Rakhlin, A. and Sridharan, K.
\newblock Online learning with predictable sequences.
\newblock arXiv:1208.3728, 2012.

\bibitem[Ravikumar et~al.(2010)Ravikumar, Wainwright, and
  Lafferty]{RavWaiLaf10}
Ravikumar, P., Wainwright, M.~J., and Lafferty, J.~D.
\newblock High-dimenstional {Ising} model selection using $\ell_1$-regularized
  logistic regression.
\newblock \emph{Annals of Statistics}, 38:\penalty0 1287--1319, 2010.

\bibitem[Snijders(2001)]{Sni01}
Snijders, T. A.~B.
\newblock The statistical evaluation of social network dynamics.
\newblock \emph{Sociological Methodology}, 31(1):\penalty0 361--395, 2001.

\bibitem[Theodor \& Shaked(1996)Theodor and Shaked]{TheSha96}
Theodor, Y. and Shaked, U.
\newblock Robust discrete-time minimum-variance filtering.
\newblock \emph{IEEE Trans. Sig. Proc.}, 44(2):\penalty0 181--189, 1996.

\bibitem[Vaswani \& Lu(2010)Vaswani and Lu]{modifiedCS}
Vaswani, N. and Lu, W.
\newblock {Modified-CS}: Modifying compressive sensing for problems with
  partially known support.
\newblock \emph{IEEE Trans. Sig. Proc.}, 58:\penalty0 4595--4607, 2010.

\bibitem[Xiao(2010)]{xiao}
Xiao, L.
\newblock Dual averaging methods for regularized stochastic learning and online
  optimization.
\newblock \emph{J. Mach. Learn. Res.}, 11:\penalty0 2543--2596, 2010.

\bibitem[Xie et~al.(1994)Xie, Soh, and de~Souza]{XieSoh94}
Xie, L., Soh, Y.~C., and de~Souza, C.~E.
\newblock Robust {Kalman} filtering for uncertain discrete-time systems.
\newblock \emph{IEEE Trans. Autom. Control}, 39:\penalty0 1310--1314, 1994.

\bibitem[Zinkevich(2003)]{Zin03}
Zinkevich, M.
\newblock Online convex programming and generalized infinitesimal gradient
  descent.
\newblock In \emph{Proc. Int. Conf. on Machine Learning (ICML)}, pp.\
  928--936, 2003.

\end{thebibliography}
\bibliographystyle{icml2013}

\end{document}